%% file: main.tex
\documentclass[pdflatex,sn-mathphys,Numbered]{sn-jnl}

\usepackage{graphicx}%
\usepackage{amsmath,amssymb,amsfonts}%
\usepackage[title]{appendix}%
\usepackage{xcolor}%
\usepackage{manyfoot}%
\usepackage{textcomp}%
\usepackage{booktabs}%
\usepackage{array}%
\usepackage{tabularx}%
\usepackage{xurl}%
\usepackage{verbatim}

\AtBeginDocument{}

\begin{document}

\title[Scout: Token, Energy, and Water Awareness]{Scout: A Game-Based Framework for Teaching Token, Energy, and Water Awareness in AI-Assisted Programming}

\author*[1]{\fnm{Abdur R.} \sur{Shahid}}\email{shahid@cs.siu.edu}

\affil*[1]{\orgdiv{School of Computing}, \orgname{Southern Illinois University Carbondale}, \orgaddress{\city{Carbondale}, \postcode{62901}, \state{IL}, \country{USA}}}

\author[2]{\fnm{Ahmed} \sur{Imteaj}}\email{aimteaj@fau.edu}

\affil[2]{\orgdiv{Electrical Engineering and Computer Science}, \orgname{Florida Atlantic University}, \orgaddress{\city{Boca Raton}, \postcode{33431}, \state{FL}, \country{USA}}}

\abstract{Programming is the largest and one of the most token-intensive uses of coding assistants, yet their interfaces conceal the tokens, energy, and water behind each request, and introductory courses do not teach this cost. Resource-awareness instruction must make these requirements understandable without discouraging productive help-seeking. This paper develops Scout, a game-based instructional framework that connects programming practice with the interpretation of resource demand. A design synthesis of self-regulated learning, formative feedback, and conjecture mapping yields six instructional principles. A formal scenario model represents complete interaction histories, separates energy from water, and preserves incomplete information. Analytical cases, based on declared teaching assumptions, establish conditions under which workflow rankings reverse, equal-energy alternatives differ in attributed water, and incomplete histories prevent defensible comparisons. A working prototype, played in one supervised sitting of an introductory programming course, implements ten missions with resource briefings, a graduated coach, five comparison challenges, evidence-labelled records, GPU-metered and token-accounted requests to a local open-weight model, and resource and control conditions. The contributions are a design rationale, a tested scenario model, an instrumented prototype, and a construct-aligned validation strategy. Empirical effectiveness remains to be established through expert appraisal and student studies.}


\maketitle

\input{body}

\backmatter

\bmhead{Acknowledgements}
OpenAI ChatGPT and Codex, and Anthropic Claude, assisted literature discovery, synthesis, drafting, figure preparation, and prototype development. The authors reviewed the final manuscript.

\input{appendix}

\bibliography{references}

\end{document}

%% file: body.tex
\section{Introduction}

Coding large language models (LLMs) have changed the conditions under which students obtain programming assistance \cite{finnie2022,becker2023,prather2023,denny2024}. Explanations, debugging suggestions, and complete implementations can be requested during the same problem-solving session, and instructors have responded with policies ranging from prohibition to deliberate integration \cite{lau2023}. Educational coding assistants have responded by introducing guidance policies that emphasize conceptual explanations and incremental support \cite{r3,r4,r5}. These developments create opportunities for timely feedback \cite{leinonen2023,hellas2023}, but also raise questions about how students decide when to request assistance, how unequal benefits accrue to students with different metacognitive skills \cite{kazemitabaar2023,prather2024}, and how students interpret the consequences of those decisions.

The resource requirements of assistance are largely absent from the student's immediate experience. A response appears in an editor or conversation window, while computation occurs on a local device or shared infrastructure. The computational cost of modern language models has become a recognized sustainability concern \cite{strubell2019,schwartz2020,devries2023}, and electricity demand from data centers is projected to grow substantially \cite{masanet2020,shehabi2024}. Energy demand depends on the workload and its execution conditions \cite{r1,samsi2023}, and water accounting introduces additional geographic and facility boundaries \cite{r2,mytton2021,siddik2021}. Neither a visible message nor a prompt count fully represents that activity. A learning environment concerned with resource awareness must therefore connect the student's interaction with an interpretable account of the underlying workflow.

This connection presents an instructional tension. Additional assistance can be educationally valuable, particularly when it resolves a misconception or enables a student to make progress. A resource display that implicitly rewards fewer requests may instead encourage avoidance of necessary help, a well-documented failure mode in interactive learning environments \cite{aleven2003}. Conversely, a display that attaches precise numbers to opaque infrastructure may create confidence without understanding. Resource-aware programming education consequently requires more than a running tally of prompts or watt-hours: it requires opportunities to compare alternatives, identify missing evidence, and justify decisions in relation to both learning and resource demand.

\subsection{Why Tokens, and Why Coding?}

Tokens are the unit through which language-model assistance is both computed and metered. Every request is processed as a sequence of prompt tokens and produces a sequence of output tokens; commercial services price usage per token, and local serving engines report the same counts. Inference energy tracks these counts closely: workload-based models fitted to measured runs explain more than 96\% of the variance in energy from input and output token counts, with output tokens having the larger effect \cite{wilkins2024}, and energy per prompt differs by more than an order of magnitude across models and prompt lengths \cite{jegham2025}. Some services also generate internal reasoning tokens that are computed and billed but not shown to the user \cite{openai2025reasoning}. A token count is therefore not itself an energy or water quantity, because the conversion depends on the model, hardware, and facility, but it is the closest learner-visible trace of the computation that a request triggers.

Coding is where this trace grows fastest. Software-related work forms the largest category of observed use of a general-purpose assistant, 37.2\% of sampled conversations in one large-scale analysis \cite{handa2025}, and 84\% of respondents to a 2025 developer survey used or planned to use AI tools, including most respondents who were learning to code \cite{so2025}. Programming requests are also unusually token-intensive. A prompt typically carries source code, error messages, and test output in addition to the question; debugging proceeds in loops of request, execution, and revision; and a learner can regenerate an answer or request a complete solution with no visible consequence. In agentic coding, where a tool reads a repository and iterates autonomously, one recent analysis found tasks consuming roughly three orders of magnitude more tokens than code chat, total tokens varying by up to 30-fold across repeated runs of the same task, input context rather than generated text dominating the cost, and accuracy saturating beyond intermediate cost \cite{bai2026}. The help-seeking habits that drive these patterns begin when students first learn to program.

Programming education has not yet treated this cost as course content. Introductory courses teach the time and space complexity of the programs that students write, but not the computational cost of the assistance used to write them. Sustainability remains weakly integrated into computing education more broadly, and employers report difficulty finding graduates who can reason about it \cite{moreira2024}. Research on generative AI in programming education has concentrated on learning gains, overreliance, academic integrity, and equity \cite{prather2023,denny2024,kazemitabaar2023,prather2024}. Introductory programming therefore offers a point of high leverage: it reaches large cohorts, precedes professional habits, and already asks students to reason about cost, correctness, and evidence. Scout targets this point by making token counts, energy, and water visible at the moment a student chooses how to seek help, while treating a larger request as legitimate when it supports learning.

\subsection{Research Gap and Motivation}

Relevant research addresses different parts of this problem. Guided coding assistants investigate how to support students without routinely revealing complete solutions \cite{r3,r4,r5}. Resource-feedback interfaces investigate awareness and selected usage responses \cite{r6}, building on a longer tradition of eco-feedback design \cite{froehlich2010,abrahamse2005}. Sustainability has entered computing curricula, but mostly through software-engineering and ethics content rather than through the act of seeking programming help \cite{peters2024,becker2015}, and none of these streams treats the token, the unit through which assistance is computed and metered, as an object of learning. Serious-game research examines learning and motivation under different instructional arrangements \cite{r7,connolly2012,clark2016}, while measurement methods characterize energy and water within specified technical boundaries \cite{r1,r2,r8,r9,henderson2020}. These streams provide complementary foundations, but do not by themselves specify how resource interpretation should be integrated into the cycle of learning to program.

The gap addressed here is the absence of an explicit design account linking four elements: programming activity, assistance decisions, interpretable resource representations, and independently assessable educational outcomes. This is a design gap identified through a focused synthesis of the selected literature, rather than a claim that no related platform exists. Its significance lies in the possibility that a technically correct resource interface can still teach an incorrect decision rule, such as equating fewer prompts with better learning or equal energy with equal water.

A game provides a suitable setting for examining these relationships because programming actions can have visible consequences and resource comparisons can be embedded within meaningful decisions. Its educational value depends on those relationships, rather than on narrative or visual polish alone. Scout uses the restoration and operation of a rover to connect introductory Python concepts with action, feedback, and reflection.

\subsection{Aim, Objectives, and Research Questions}

The aim of this study is to develop a theoretically grounded and technically interpretable framework for teaching resource-aware help-seeking during LLM-assisted programming. Resource awareness is treated as a capacity to interpret and justify a decision, rather than as a requirement to minimize assistance.

Four objectives guide the work. O1 is to derive instructional requirements from research on guided programming assistance, self-regulation, feedback, and resource communication. O2 is to define a scenario model that represents complete workflows, compatible energy-and-water boundaries, and missing information. O3 is to instantiate the requirements in Scout's mission structure and specify the responsibilities of browser and institutional services. O4 is to establish an evaluation framework that distinguishes resource literacy, help-seeking, usability, and independent programming learning.

These objectives motivate three research questions. RQ1 asks which design principles can connect resource explanations with productive programming activity. RQ2 asks how scenario representations can expose workflow trade-offs and uncertainty without implying unsupported measurements. RQ3 asks what observable evidence would support or challenge the educational mechanisms embodied in the design. The present paper addresses these questions through design synthesis, formal specification, and analytical demonstration. Estimating educational effects is reserved for the empirical validation stage.

\subsection{Contributions}

The first contribution is a traceable instructional framework comprising six design principles and a conjecture map. Each principle links a design decision to a proposed learning process and specifies observations that could contradict the intended mechanism. The contribution is the integrated account of these relationships, rather than the invention of self-regulation, feedback, or hint-based tutoring.

The second contribution is an operation-based scenario model with explicit system boundaries, water-accounting definitions, missing-data semantics, and interval-based comparison rules. Its analytical cases demonstrate how seemingly intuitive resource rankings can reverse. The underlying physical equations and comparison relations are established tools; their contribution here is their use as inspectable instructional content.

The third contribution is a Scout instantiation and modular reference architecture that separates learning interaction from resource scenarios and optional live services. A working local prototype implements the missions with resource briefings, a graduated assistance policy, the scenario engine, evidence-labelled resource records, GPU-metered and token-accounted assistance requests, and support for comparative studies; a zero-installation in-browser runtime and institutional services remain specified extensions. This distinction prevents architectural possibilities from being presented as deployed capabilities.

The fourth contribution is a construct-aligned validation framework, including candidate resource-literacy tasks, an explanation rubric, a formative evaluation strategy, and a controlled-comparison design. Together, these contributions make the proposed instructional mechanisms and the limits of the current evidence explicit. Table~\ref{tab:objectives} maps the objectives to the resulting artifacts and their evidence status.

\begin{table}[!t]\caption{Objectives, contributions, and evidence status.\label{tab:objectives}}\centering\footnotesize\begin{tabularx}{\textwidth}{>{\raggedright\arraybackslash}p{.22\textwidth}XX}\toprule
\textbf{Objective} & \textbf{Research output} & \textbf{Evidence status}\\
\midrule
O1: Instructional requirements & Six principles and conjecture map (C1) & Theory-informed design synthesis\\
O2: Scenario representation & Formal model and analytical cases (C2) & Specification and reproducible calculations\\
O3: Scout instantiation & Mission design, prototype, and reference architecture (C3) & Implemented and automatically tested (Table~\ref{tab:status}); measurement validation on study hardware pending; in-browser runtime and institutional services specified\\
O4: Educational validation & Constructs, tasks, rubric, and study design (C4) & Evaluation framework; participant studies remain future work\\
\bottomrule\end{tabularx}\end{table}

\section{Related Work}

Table~\ref{tab:positioning} summarizes the five research streams on which the framework draws, the contribution each stream makes, and the integration gap that motivates Scout. The subsections below discuss each stream in turn.

\begin{table}[!t]\caption{Research streams informing the framework and the integration gap addressed by Scout.\label{tab:positioning}}\centering\footnotesize\begin{tabularx}{\textwidth}{>{\raggedright\arraybackslash}p{.17\textwidth}>{\raggedright\arraybackslash}p{.17\textwidth}>{\raggedright\arraybackslash}X>{\raggedright\arraybackslash}X}\toprule
\textbf{Stream} & \textbf{Representative work} & \textbf{Established contribution} & \textbf{Gap relative to this study}\\
\midrule
Generative AI in programming education & \cite{finnie2022,becker2023,prather2023,denny2024,kazemitabaar2023,prather2024} & Capabilities, risks of over-reliance, instructor responses & Resource consequences of assistance are not represented\\
Guided coding assistants & \cite{r3,r4,r5,hellas2023,leinonen2023} & Guardrailed, conceptual, or incremental help policies & No link between help policy and complete-workflow resource demand\\
Help-seeking and self-regulation & \cite{aleven2003,roll2011,r12,panadero2017,loksa2016,tankelevitch2024} & Help avoidance and abuse; metacognitive scaffolds & Resource information as an input to help decisions is unexamined\\
Eco-feedback and sustainable computing & \cite{froehlich2010,abrahamse2005,r6,peters2024,pinto2017,pereira2021} & Awareness interfaces; sustainability in curricula and software & Awareness and behavior rarely separated from learning outcomes\\
AI energy and water accounting & \cite{r1,r2,strubell2019,henderson2020,luccioni2023bloom,samsi2023,mytton2021} & Boundary-specific measurement and estimation methods & Accounting distinctions rarely treated as instructional content\\
\bottomrule\end{tabularx}\end{table}

\subsection{Generative AI in Programming Education}

Early evaluations showed that code-generating models could solve typical introductory exercises with high accuracy \cite{finnie2022}, prompting analyses of the opportunities and risks for computing education \cite{becker2023,prather2023,denny2024}. Models can generate exercises and explanations \cite{sarsa2022}, enhance error messages \cite{leinonen2023}, and respond to novice help requests, although response quality varies and errors are not always detected by learners \cite{hellas2023}. Controlled and observational studies indicate that access to code generators can improve task completion without necessarily impairing later performance \cite{kazemitabaar2023}, but also that benefits depend on metacognitive skill, so that the same tool may widen gaps between students \cite{prather2024}. Instructors' planned responses range from restriction to deliberate integration \cite{lau2023}. These studies establish that assistance design and students' regulation of assistance matter educationally; they do not address the resource consequences of the help-seeking decisions they study.

\subsection{Guided Coding Assistance}

CodeAid provides conceptual explanations, pseudocode, and debugging suggestions within a classroom programming assistant \cite{r3}. CodeHelp describes a guardrailed prompting pipeline intended to support learners without directly supplying complete solutions \cite{r4}. Both provide relevant precedents for assistance that leaves meaningful programming work with the student. They do not establish a universal ordering between hints and worked solutions, nor do they demonstrate that restricted output necessarily reduces resource demand.

The CS50 tutor research illustrates the need to evaluate instructional behavior as prompts and models change \cite{r5}. A response may be technically correct yet provide too much of the solution, or may preserve the answer while offering insufficient guidance. Assistance should therefore be represented as an instructional policy with explicit expectations about relevance, disclosure, and the learner's subsequent action, consistent with research on hint design in intelligent tutoring systems \cite{aleven2003,roll2011}.

This distinction matters for resource instruction. The visible response is not necessarily the complete workload: generation, checking, or rewriting can occur before a short hint is displayed. A comparison that counts only visible messages can consequently mischaracterize the work performed. Scout's scenario model represents those internal operations when they are part of the declared scenario.

\subsection{Help-Seeking and Self-Regulation}

Help-seeking research distinguishes productive help use from help avoidance and help abuse, such as rapidly requesting hints until a solution is revealed \cite{aleven2003}. Metacognitive feedback on help-seeking behavior can improve help-seeking skill in tutoring environments, although transfer to domain learning is not automatic \cite{roll2011}. In programming, explicit guidance on problem-solving stages and self-monitoring improves productivity and self-efficacy \cite{loksa2016}. Generative AI places new metacognitive demands on users, who must formulate requests, evaluate outputs, and decide how much to rely on them \cite{tankelevitch2024}. These findings suggest that resource information should be introduced as one further input to a regulated help-seeking decision, rather than as an external constraint that substitutes for that decision.

\subsection{Resource Feedback and Resource Literacy}

GPTFootprint introduced resource feedback into ChatGPT interactions and reported increased awareness alongside limited success in reducing use in an exploratory study \cite{r6}. The findings demonstrate why awareness and behavior should be assessed separately. Eco-feedback research in household energy reaches a similar conclusion: feedback can support conservation, but its effects depend on framing, granularity, and the actionability of the information presented \cite{froehlich2010,abrahamse2005}. Their implications for programming education remain conditional on the instructional context, the available assistance, and the reasons students seek help.

For this study, resource literacy has three components: interpreting a resource quantity, evaluating whether a comparison is supported, and justifying a decision under stated conditions. Concern, confidence, and willingness to reduce use are related but distinct constructs. A student may accurately understand a resource trade-off and reasonably choose additional assistance. The framework therefore evaluates the quality of reasoning rather than conformity to a predetermined low-use response.

Sustainability is increasingly treated as a computing-education objective, but a systematic review found that coverage concentrates on general awareness, software-engineering practice, and ethics \cite{peters2024}. Software-engineering research has articulated sustainability as a design concern \cite{becker2015} and shown that energy efficiency depends on implementation choices, including the programming language \cite{pinto2017,pereira2021}. These lines of work treat the student as a future producer of efficient software. Scout addresses a complementary and more immediate role: the student as a present consumer of computationally intensive assistance.

\subsection{Game-Based Learning and Feedback}

Wouters and colleagues distinguish learning, retention, and motivation in their meta-analysis of serious games \cite{r7}. Their findings, together with subsequent systematic reviews \cite{connolly2012,clark2016}, indicate that the instructional arrangement matters and do not support treating a game format as sufficient evidence of educational value. Reviews of serious games for programming similarly report a wide range of designs, but limited evaluation of learning outcomes beyond the game itself \cite{miljanovic2018}. Constructionist accounts emphasize that games are most productive when learners act on and reason about a system rather than merely receive rewards \cite{kafai2015}. For Scout, the relevant mechanism is the connection between code, system behavior, and revision: a program changes a rover capability, and test feedback makes the consequences inspectable.

Self-regulated learning provides a structure for this cycle. Forethought includes interpreting a task and selecting a strategy; performance includes implementation, monitoring, and help-seeking; reflection includes evaluating the outcome and revising the strategy \cite{r12,panadero2017}. Formative feedback supports this process when it provides information that can guide the next action, with its usefulness depending on the task and learner \cite{r13,hattie2007}. Automated feedback on programming exercises has been studied extensively, with most systems focusing on correctness and fewer on strategy or process \cite{keuning2018}. Resource explanations are integrated into this cycle rather than appended as unrelated warnings.

\subsection{Energy and Water Accounting}

Energy measurement depends on a defined hardware boundary. CodeCarbon documents the use of energy counters and power-based alternatives, including their different measurement properties \cite{r8}. Standardized reporting frameworks and calculators for machine-learning energy and emissions emphasize the declaration of hardware, location, and boundary \cite{henderson2020,lacoste2019,lannelongue2021}. Training-focused studies first exposed the scale of these costs \cite{strubell2019,patterson2021}, while life-cycle and system-level analyses showed that inference, idle capacity, and embodied impacts can be substantial \cite{luccioni2023bloom,wu2022}. Inference benchmarks report energy that varies with model size, hardware, batch configuration, and output length \cite{samsi2023,jegham2025}. Work on AI deployment energy further motivates reporting workload and execution conditions instead of assuming a universal per-message factor \cite{r1}. Water introduces a separate set of distinctions among site cooling, electricity-related use, consumption, and withdrawal \cite{r2,r9,mytton2021}. Electricity-related water depends on the generation mix \cite{macknick2012}, and data-center footprints vary geographically with both cooling technology and local water stress \cite{siddik2021}.

These distinctions constrain the design of educational scenarios. A physically correct equation with assumed parameters remains an estimate. A component measurement does not automatically represent an entire device. A facility-average allocation does not establish the marginal water consequence of a particular request. The framework makes these qualifications objects of instruction, with unfamiliar comparisons used to assess whether learners can apply them.

\section{Methodological Approach}\label{sec:method}

\subsection{Conceptual Development and Design Synthesis}

The work follows a design-science-informed approach in which a relevant problem is translated into design objectives, an artifact, demonstration, and evaluation \cite{r10,hevner2004}. It also draws on design-based research in education, which treats an intervention as an embodiment of testable conjectures about learning \cite{dbrc2003}. Following Gregor and Hevner \cite{gregor2013}, the contribution is positioned as an improvement: a new solution design for a recognized problem, with knowledge claims limited to the level of evidence presented. The present contribution comprises the conceptual framework, a system specification grounded in an existing prototype, and analytical validation of the scenario relationships. Participant evaluation is a subsequent stage.

The evidence base is a focused synthesis of primary studies, established educational methods, and technical accounting sources. Sources were selected for direct relevance to guided programming, resource feedback, game learning, self-regulation, design evaluation, or physical accounting. This selection supports a bounded design argument; it does not constitute a systematic review or establish exhaustive coverage of related platforms.

The synthesis was organized around four tensions: assistance versus independent application; visibility versus interpretability; individual interaction versus complete workflow; and actionable choice versus infrastructure uncertainty. The six principles in Section~\ref{sec:framework} respond to these tensions. Their justification is recorded in the principle-to-evidence matrix, allowing subsequent evaluations to revise a principle rather than merely add interface features.

\subsection{Conjecture Mapping and Claim Boundaries}

\begin{figure}[!t]\centering
\includegraphics[width=\textwidth]{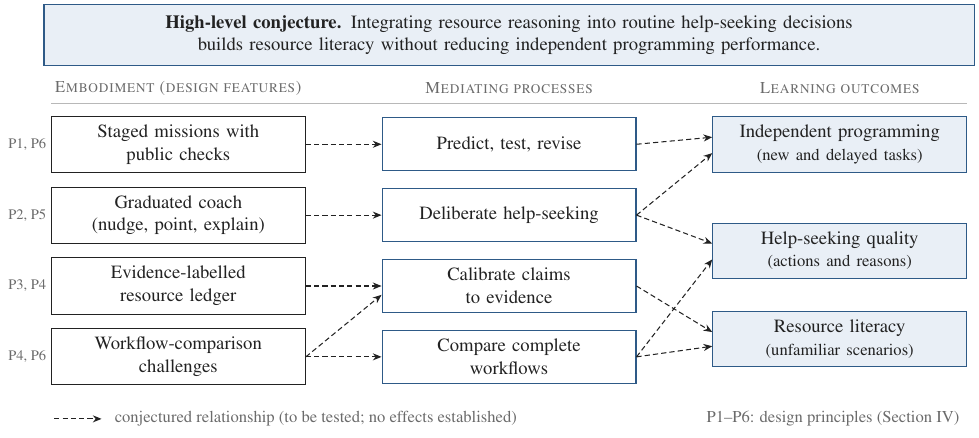}
\caption{Instructional conjecture map, following Sandoval \cite{r14}, linking Scout design features to proposed mediating processes and separately assessed learning outcomes. Tags P1--P6 indicate the design principles (Section~\ref{sec:framework}) that each feature embodies. Dashed arrows denote conjectures to be investigated empirically; no effects are established by this paper.\label{fig:conjecture}}\end{figure}

Conjecture mapping distinguishes the features of a learning environment from the processes and outcomes those features are expected to produce \cite{r14}. Figure~\ref{fig:conjecture} applies this distinction to Scout. For example, displaying a session history is a design feature; comparing attempts is a proposed process; interpreting an unfamiliar workflow is an assessable outcome. Observing the feature alone does not establish either the process or the outcome.

The evaluation strategy also distinguishes formative and summative purposes and controlled and naturalistic settings, following the logic of the Framework for Evaluation in Design Science \cite{r11}. Arithmetic checks establish internal scenario consistency. Expert review can establish the presence of identified content or usability problems. Student observations can reveal how the design is interpreted. Causal claims about learning require a suitable comparative design. These distinctions determine the scope of the paper's conclusions.

\section{Instructional Framework}\label{sec:framework}

\subsection{Staged Programming and Resource Concepts}

\begin{figure}[!t]\centering
\includegraphics[width=\textwidth]{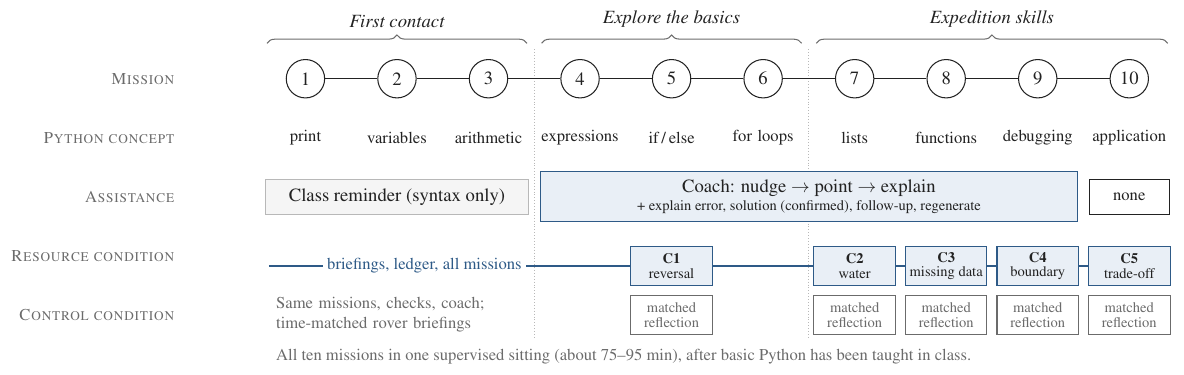}
\caption{Mission sequence in the Scout prototype. Ten missions in three phases introduce one Python concept each. Assistance moves from a syntax-only class reminder (missions 1--3) to a graduated coach (4--9) and is withdrawn for the independent mission (10). All ten missions are completed in one supervised sitting after the corresponding Python constructs have been taught in class. In the resource condition each mission opens with a short energy-and-water briefing, an evidence-labelled ledger is visible throughout, and five comparison challenges follow missions 5 and 7--10: C1 workflow-length reversal, C2 equal energy with unequal water, C3 missing data, C4 boundary mismatch, C5 trade-off. The control condition receives rover-engineering briefings of the same length and time-matched reflection prompts at the same points.\label{fig:progression}}\end{figure}

The first principle (P1) is progressive integration. Introductory missions isolate printing, variables, and simple expressions before combining decisions, repetition, lists, and functions. Resource concepts follow a parallel progression: identifying energy and water as separate quantities; interpreting a scenario value; comparing complete workflows; and recognizing uncertainty. This arrangement limits the number of unfamiliar demands introduced simultaneously, consistent with cognitive-load accounts of novice problem solving \cite{sweller1988}. Figure~\ref{fig:progression} shows the parallel progression across mission stages and its relation to the self-regulated learning cycle.

The second principle (P2) is actionable assistance. A student should be able to relate a hint or explanation to a specific difficulty and make a subsequent code change or test. The relevant educational process is deliberate help-seeking, as distinct from help avoidance and help abuse \cite{aleven2003,roll2011}. Additional support is appropriate when it enables understanding or progress; an assistance policy should not designate the shortest response as inherently preferable. Scout operationalizes this principle as graduated help in the tradition of intelligent tutoring systems \cite{aleven2003}: the learner chooses a \emph{nudge} (a diagnostic question), a \emph{pointer} to the relevant line or construct, or an \emph{explanation} of the concept using a different example. After a failed run the learner can also ask for an explanation of the error, and a complete worked solution is available only after an explicit confirmation that offers a hint first. Each response supports a follow-up question in the same thread and can be regenerated; both are recorded because they add processing to the workflow. The chosen action, the preceding attempts, and the subsequent revision are recorded, so that help-seeking can be analyzed as a sequence rather than a count. Because model-generated hints may disclose too much, generated text passes through an answer guard that removes code blocks and lines reproducing the reference solution, and withholds the hint when too much would remain.

The third principle (P3) is workflow visibility. A resource comparison includes the sequence of attempts, model interactions, and declared internal processing required to reach an outcome. This representation encourages learners to examine the relationship between an assistance request and the work that follows. It also exposes cases in which a superficially brief interaction requires substantial additional activity.

\subsection{Epistemic Transparency and Learner Agency}

The fourth principle (P4) is epistemic transparency. Each resource quantity has a mode, boundary, provenance, and completeness status. Each value carries one of four evidence labels: \emph{measured} (a hardware energy counter), \emph{estimated} (a calibrated average power multiplied by elapsed time), \emph{authored} (a declared teaching range multiplied by elapsed time, reported as an interval), or \emph{unknown}. Missing information produces an incomplete comparison rather than a zero value. The intended process is calibration: learners recognize the strength and limits of the evidence supporting a decision.

The fifth principle (P5) is nonpunitive agency. Resource information informs a decision but does not remove necessary assistance or penalize accommodations. Scout does not rank students by prompts, energy, or water. Such rankings would confound preparation, task difficulty, and support needs with a narrow resource indicator, and could amplify the unequal benefits of AI assistance already observed among novices \cite{prather2024}. Reflection instead asks students to explain why an action was useful and what could be inferred from its scenario profile.

The sixth principle (P6) is transfer-oriented assessment. Supported task performance, resource interpretation, and independent programming are evaluated separately. An unfamiliar resource scenario tests whether the student can apply a distinction outside the game's wording; a new unassisted coding task examines application beyond the supported exercise. A delayed assessment is required when the claim concerns retention.

Table~\ref{tab:principles} summarizes the principles and the evidence that would challenge their proposed mechanisms. The framework thus specifies failure conditions as well as intended effects. A student who avoids help because of the display, or selects a lower prompt count despite contrary workflow evidence, provides a diagnostic observation that the corresponding principle has not been successfully realized.

\begin{table}[!t]\caption{Design principles, proposed mechanisms, and observations that would challenge them.\label{tab:principles}}\centering\footnotesize\begin{tabularx}{\textwidth}{>{\raggedright\arraybackslash}p{.22\textwidth}XX}\toprule
\textbf{Principle} & \textbf{Proposed mechanism} & \textbf{Disconfirming observation}\\
\midrule
P1 Progressive integration & Coordinate programming and resource concepts & Correct interface selections without explanations\\
P2 Actionable assistance & Connect support to a subsequent revision or test & Copying responses or avoiding necessary help\\
P3 Workflow visibility & Compare complete histories & Ranking by prompt count despite contrary workflow evidence\\
P4 Epistemic transparency & Calibrate interpretation to evidence & Treating missing values as zero or estimates as measurements\\
P5 Nonpunitive agency & Consider resources while retaining access to help & Guilt, exclusion, or strategic help avoidance\\
P6 Transfer-oriented assessment & Apply concepts beyond familiar cues & In-game success without unfamiliar-task performance\\
\bottomrule\end{tabularx}\end{table}

\section{Scout Instantiation and System Architecture}\label{sec:scout}

\subsection{Mission Structure and Interaction}

Scout is placed within an introductory programming course rather than replacing its Python instruction. Students complete all ten missions in one supervised sitting of approximately 75--95 minutes, scheduled after the course has introduced output, variables, arithmetic, selection, iteration, lists, and functions. Python is therefore the medium of the activity, and resource reasoning is its instructional content. The missions are framed as an expedition in which a rover progressively gains operational capabilities. The student establishes communication, assigns an identity, calculates available power, interprets weather, handles a warning, controls movement, inspects readings, constructs a reusable function, repairs a collector, and completes an independent application. Each task links a Python concept to a visible consequence in the game world.

The learning cycle comprises task interpretation, implementation, execution, feedback inspection, revision, and explanation. Each mission opens with a briefing of two to three minutes that presents one resource idea, a worked numerical example, a small calculator (energy, water, data-center water, or repetition), and an ungraded quick check whose first answer is recorded; a glossary defines the eighteen terms used in the briefings. The first missions provide a one-line syntax reminder and a short path to successful execution. Later activities introduce assistance choices and comparisons of complete histories. Reflection is placed after relevant experience rather than used as a lengthy entry requirement before the first code attempt.

\begin{figure}[!t]\centering
\includegraphics[width=\textwidth]{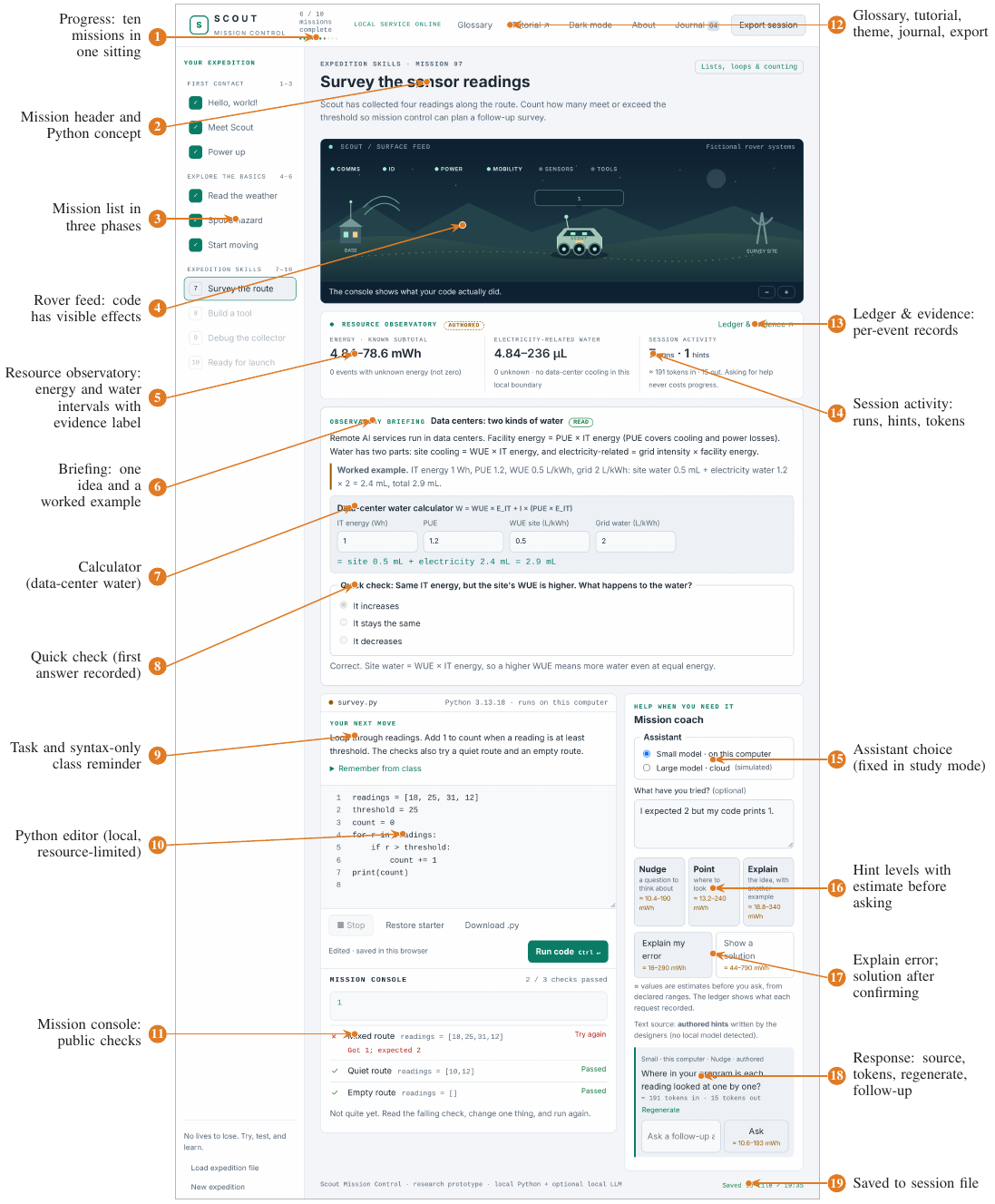}
\caption{Scout student interface in the resource condition (mission 7, after a failed check and one Nudge request). Numbered callouts mark the main components. The screenshot was taken without a local model or GPU, so the coach shows an authored hint and resource values carry the \emph{authored} label; on the study workstation, coach requests are labelled \emph{measured}.\label{fig:interface}}\end{figure}

The current prototype implements this sequence as a locally hosted web application (Table~\ref{tab:status}). Student code runs in a resource-limited child process on the learner's computer and is evaluated against public checks held by the host rather than the browser. Coach requests are metered on the study workstation's GPU (Section~\ref{sec:metering}). Because CPU energy counters are often unreadable without administrator access, code executions, and coach requests on computers without a supported GPU, receive an authored interval when neither a measurement nor a calibration is available, and are labelled accordingly. The same comparison activities can therefore be delivered on any computer without live resource telemetry. Figure~\ref{fig:interface} shows the student interface during a mission.

\begin{table}[!t]\caption{Implementation status of the Scout prototype (version 5.4). ``Tested'' refers to automated software tests, not to evaluation with learners.\label{tab:status}}\centering\footnotesize\begin{tabularx}{\textwidth}{>{\raggedright\arraybackslash}p{.24\textwidth}>{\raggedright\arraybackslash}X>{\raggedright\arraybackslash}p{.2\textwidth}}\toprule
\textbf{Component} & \textbf{Implementation} & \textbf{Status}\\
\midrule
Missions and checks & Ten missions; public checks with input substitution; reference solutions held server-side & Implemented, tested\\
Python execution & Local child process with CPU, memory, and output limits; per-run stop; not a security sandbox & Implemented, tested\\
Briefings & Ten resource briefings with worked examples, calculators, and quick checks; glossary; length-matched control briefings & Implemented, end-to-end tested\\
Assistance & Class reminder (missions 1--3); coach (4--9) with three hint levels, error explanation, confirmed solution, follow-up, and regeneration, using a local open model or labelled authored hints; answer guard; none on mission 10 & Implemented, tested with a mock model; exercised with a local 7B model\\
Scenario engine & Equations~(\ref{eq:session})--(\ref{eq:dominance}); five comparison states; five authored challenges & Implemented, unit-tested\\
Resource records & Measured, estimated, authored, or unknown energy and electricity-related water per event; prompt and output tokens and model timings per request & Implemented, tested\\
Inference metering & GPU board power sampled at 100~ms with idle-baseline subtraction; token-based estimate calibrated against measured requests & Implemented, tested with simulated sensor; validation on study hardware pending\\
Study support & Resource and control conditions; pseudonymous identifiers; consent gate; automatic per-session files; analysis tables & Implemented, end-to-end tested\\
Accessibility & Keyboard operation, live regions, reduced motion; automated WCAG~2 A/AA audit & No violations detected\\
In-browser runtime & Pyodide worker for zero-installation use & Specified\\
Institutional services & Authentication, isolated execution, instructor views & Specified\\
\bottomrule\end{tabularx}\end{table}

\subsection{Authored and Live Assistance}

Assistance can be supplied through versioned authored responses or through a live model. Authored responses make the instructional sequence reproducible across participants and allow the prototype to run without a model. A planned extension will also execute Python in the browser; Pyodide provides a documented option for running Python in a browser worker \cite{r15}, which would remove the local installation step.

An authored response is visibly identified as such. Its scenario resource profile represents a teaching case, not a model call that actually occurred. Live-service mode instead sends a request to a local or institutional model; the prototype uses a locally served open-weight coding model through the Ollama interface \cite{r16}. For a study, a single model version (for example, a 32-billion-parameter coding model on one workstation GPU) is fixed for all participants, and its name and content digest are stored with every request. The model, prompt policy, generated response, and internal operations then become part of the observed record. These modes are distinct experimental conditions and should not be silently combined.

A session record includes task and scenario versions, code revisions, executions, outcomes, assistance requests and responses, and explanations. The resource interpretation is stored alongside the operation history so that a later analysis can reconstruct what information was available when the student chose an action. Identifiers used for research should be pseudonymous, with consent and retention rules established before participant collection, in line with ethical frameworks for learning analytics \cite{slade2013}. The prototype saves every session automatically as two local files, a complete session snapshot and an append-only event log, and also offers a browser export; every event carries a sequence number so that duplicate records from the two sources are merged during analysis. Sessions interrupted by a closed browser can be resumed from the saved snapshot.

\subsection{Measured Inference Energy and Token Accounting}\label{sec:metering}

On the study workstation each coach request is metered. A sampler reads board power for all NVIDIA GPUs through \texttt{nvidia-smi} every 100~ms while the request runs and integrates the readings, $E_{\mathrm{gross}}=\sum_i P_i\,\Delta t_i$. At start-up the host averages idle GPU power $P_{\mathrm{idle}}$ over 10~s, and each request is charged the energy above idle, $E_{\mathrm{net}}=E_{\mathrm{gross}}-P_{\mathrm{idle}}\,t$; both values are stored. The model is kept resident and warmed before the session, and the model-loading duration reported by the serving engine is stored with every request so that requests that include loading can be identified. These records carry the label \emph{measured} and the boundary \emph{GPU boards}; processor, memory, storage, and power-supply losses are outside the boundary, and a whole-computer figure requires a wall-plug measurement of the same request set. Water for a measured request is the measured energy multiplied by a grid water-intensity coefficient that the researcher records with its region, year, and source, as in Eq.~(\ref{eq:water}).

The serving engine reports the number of prompt tokens processed, the number of output tokens generated, and the durations of prompt processing and generation; these are stored with the energy record, the model name, and the model digest. Before each request the resource condition displays an estimated interval. A calibration tool sends a fixed set of requests covering every coach mission, hint level, error explanation, and solution, fits $E\approx a\,n_{\mathrm{prompt}}+b\,n_{\mathrm{output}}$ to the measured energies by least squares, and sets the displayed interval to the fitted value multiplied by the 10th and 90th percentiles of the measured-to-fitted ratios. The interface states whether a displayed estimate is derived from these measurements or from declared teaching ranges, and the ledger shows the recorded value after the response. This arrangement allows a student to compare an anticipated cost with an observed one, and allows analysis to relate help-seeking choices to both token counts and energy.

\subsection{Multiple Students and Institutional Delivery}

\begin{figure}[!t]\centering
\includegraphics[width=\textwidth]{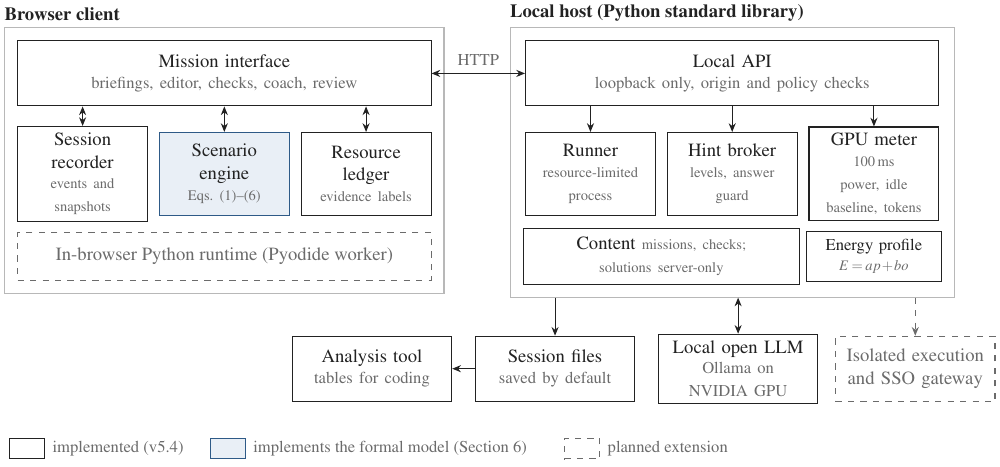}
\caption{Scout architecture. Solid outlines mark components implemented in the prototype; the shaded component implements the formal model of Section~\ref{sec:model}; dashed outlines mark planned extensions. The local open model is optional for demonstrations: without it the coach serves labelled authored hints. On the study workstation the GPU meter records each coach request, the energy profile holds the calibrated token coefficients, and every session is saved to local files.\label{fig:architecture}}\end{figure}

The reference architecture in Figure~\ref{fig:architecture} separates the learning interface, scenario engine, execution service, assistance service, and records. The current prototype operates on a single computer and saves each session to local files. An institutional deployment adds authentication, persistent per-student records, authorized instructor views, and isolated execution when student code runs on a server.

A shared model service also requires admission control, scheduling, and records of queueing, failures, and cancellation. Serving engines can change batching and memory management \cite{r17}, and measured inference energy varies with these configurations \cite{samsi2023}; therefore, prompt counts alone cannot characterize shared infrastructure. Physical per-student attribution would require additional telemetry and an explicit allocation rule. These engineering responsibilities are represented in the architecture without attributing deployment capacity to the single-user prototype.

The architecture preserves an important experimental separation: the learning activity can remain fixed while the assistance provider changes, and a resource scenario can remain fixed while interface explanations are varied. This modularity supports interpretable comparisons and avoids confounding an educational intervention with an unrelated change in model capability.

\section{Formal Scenario Model}\label{sec:model}

\subsection{Session Representation and Boundary Compatibility}\label{sec:session}

Let a learning session be an ordered sequence of operations, each associated with a task state, action, outcome, and resource description. The operation sequence includes code execution, assistance processing, response generation, and any checking or regeneration represented by the scenario. Its modeled energy is defined over a stated boundary B:

\begin{equation}
\begin{gathered}
\mathcal{S}_s=(o_1,\ldots,o_m),\qquad E_s^{(B)}=\sum_{o\in\mathcal{S}_s}E_o^{(B)},\\
E_o^{(B)}=\frac{1}{3600}\int_0^{T_o}P_o(t)\,\mathrm{d}t.
\end{gathered}
\label{eq:session}\end{equation}

Here $P_o(t)$ is the power profile assigned to operation o and $T_o$ is its duration. The equation expresses energy in watt-hours when power is in watts and time in seconds. In a teaching scenario the profile is authored; in a measurement study it would need a compatible observation source. A constant-power profile reduces to $E_o=P_oT_o/3600$.

A comparison is admissible only when the task outcome and accounting boundaries are compatible. For example, comparing one workflow's CPU-package energy with another's facility energy does not establish a like-for-like ranking; this is the same boundary-declaration principle advocated for machine-learning energy reporting \cite{henderson2020,lannelongue2021}. The representation therefore records boundary, included components, excluded components, parameter source, and data completeness as first-class attributes.

Let $K_s$ and $U_s$ denote the known and unknown operation sets. The system reports a known-component subtotal and a separate completeness indicator:

\begin{equation}
\begin{gathered}
E_s^{K}=\sum_{o\in K_s}E_o,\qquad E_s^{\mathrm{tot}}=E_s^{K}+E_s^{U},\\
E_s^{U}\ge0,\qquad \chi_s=\mathbf{1}\{U_s=\varnothing\}.
\end{gathered}
\label{eq:missing}\end{equation}

Provided that unknown operation energy is nonnegative, the known subtotal is a lower bound on total energy, not a complete estimate. If no defensible bound exists for the unknown part, a complete-workflow ranking may be indeterminate. The interface exposes this state rather than substituting a visually convenient zero.

\subsection{Water Accounting}

For electricity-related water, the relevant energy is multiplied by a location- and period-specific water intensity. When energy is expressed in watt-hours and intensity in liters per kilowatt-hour, the numerical product gives milliliters. A data-center scenario additionally distinguishes IT energy from facility electricity:

\begin{align}
W_{\mathrm{elec}}~[\mathrm{mL}]&=E_{\mathrm{elec}}~[\mathrm{Wh}]\; I_{\mathrm{grid}}~[\mathrm{L/kWh}],\label{eq:water}\\
E_{\mathrm{facility}}&=\mathrm{PUE}\;E_{\mathrm{IT}},\label{eq:pue}\\
W_{\mathrm{site}}~[\mathrm{L}]&=\mathrm{WUE}_{\mathrm{site}}~[\mathrm{L/IT\!\!-\!kWh}]\;E_{\mathrm{IT}}~[\mathrm{kWh}].\label{eq:wue}
\end{align}

PUE denotes power usage effectiveness \cite{r9}, $\mathrm{WUE}_{\mathrm{site}}$ denotes site water usage effectiveness with IT energy as its denominator, and $I_{\mathrm{grid}}$ denotes electricity-related water intensity. The formulation prevents site WUE from being incorrectly multiplied by facility energy. A source-based WUE that already includes upstream electricity water must be treated under its own boundary to avoid double counting \cite{r2,r9}.

The model records whether a coefficient refers to consumption or withdrawal and whether it is an annual allocation or an operational observation. These quantities are not interchangeable. A local air-cooled computer scenario can exclude a remote facility's cooling term while retaining electricity-related water, whose intensity depends on the generation technologies serving the grid \cite{macknick2012}. Location follows the computing service, rather than automatically following the student's browser.

\subsection{Partial Ordering and Uncertainty}

Resource alternatives need not admit a single ranking. Consider alternatives a and b with compatible outcomes and boundaries. If the upper bounds of both a's energy and water are no greater than the corresponding lower bounds for b, with at least one strict inequality, a robustly dominates b within the declared interval assumptions:

\begin{equation}
a\prec_R b\quad\Longleftarrow\quad
\overline E_a\le\underline E_b\ \land\ \overline W_a\le\underline W_b,
\label{eq:dominance}\end{equation}

This relation is deliberately conservative. If the intervals overlap, the ranking may depend on the chosen parameters; if energy and water favor different alternatives, the comparison presents a trade-off. Neither case justifies a universal ``best'' button. The intervals describe declared scenario ranges and are not statistical confidence intervals unless a probability model supports that interpretation.

\begin{figure}[!t]\centering
\includegraphics[width=\textwidth]{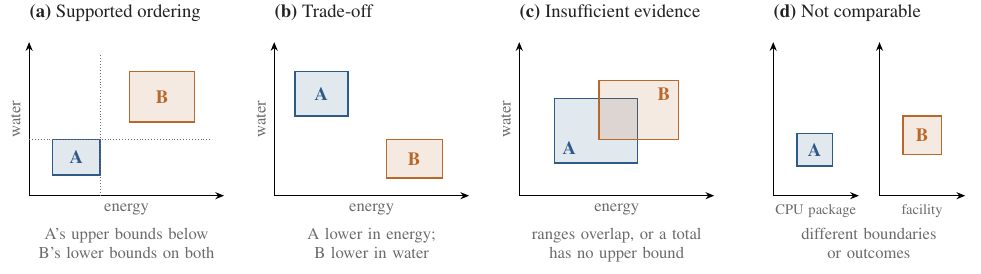}
\caption{Comparison states under the interval relation in Equation~(\ref{eq:dominance}). Rectangles show declared energy and water ranges for workflows A and B. (a) Supported ordering: A's upper bounds do not exceed B's lower bounds on either dimension. (b) Trade-off: energy and water favor different workflows. (c) Insufficient evidence: the ranges overlap, or an unknown operation leaves a total without an upper bound. (d) Not comparable: the workflows use different accounting boundaries or reach different outcomes.\label{fig:ordering}}\end{figure}

Combining this relation with the admissibility condition of Section~\ref{sec:session}, the instructional model distinguishes four comparison states, illustrated in Figure~\ref{fig:ordering}: a supported ordering (in either direction), a trade-off between dimensions, insufficient evidence for an ordering, and a comparison that is not admissible because outcomes or accounting boundaries differ. The prototype presents the supported ordering as two options, so learners choose among five responses.

 Students can be asked to identify the state before selecting an action. Educational adequacy remains a separate constraint: dominance among resource quantities does not establish equivalence in learning support.

\section{Analytical Characterization}\label{sec:analysis}

\subsection{Workflow Ranking Reversal}

The first analytical case examines whether a guided workflow necessarily requires less energy than a direct response. Let a direct response use a constant profile $P_D$ for duration $T_D$. Let each of n guided interactions use profile $P_H$ for duration $T_H$. For the operations included in the declared boundary, the guided workflow has lower modeled energy precisely when:

\begin{equation}
\begin{aligned}
E_H(n)=\frac{nP_HT_H}{3600}&<\frac{P_DT_D}{3600}=E_D\\
\Longleftrightarrow\quad n&<\frac{P_DT_D}{P_HT_H},\qquad P_HT_H>0.
\end{aligned}
\label{eq:threshold}\end{equation}

This condition makes the inference conditional on both the per-interaction profile and the number of interactions. It provides a counterexample to any universal ranking based only on whether the interface supplies a hint or a full answer.

For illustration, set $P_D$ = 60 W, $T_D$ = 20 s, $P_H$ = 15 W, and $T_H$ = 15 s. These are selected teaching assumptions. The direct workflow uses 0.333 Wh; four guided interactions use 0.250 Wh; six use 0.375 Wh. Their ordering reverses at $n=16/3$, approximately 5.33 interactions. Figure~\ref{fig:analytical}a shows this example, and Figure~\ref{fig:analytical}b shows how the crossover $n^*=P_DT_D/(P_HT_H)$ moves across a range of guided-interaction profiles. Because inference energy depends on model, hardware, and output length \cite{samsi2023,jegham2025}, any realistic crossover would need to be established for a specific deployment. The crossover characterizes the authored scenario and is not a real-model efficiency threshold.

\subsection{Equal Energy and Unequal Water}

The second case holds IT energy and facility electricity constant across two hypothetical serving windows. With 0.5 Wh of IT energy, PUE = 1.2, and $I_{\mathrm{grid}}$ = 0.4 L/kWh, electricity-related water is 0.24 mL in each window. If $\mathrm{WUE}_{\mathrm{site}}$ is 0.2 L/IT-kWh in one window and 1.0 L/IT-kWh in the other, site water is 0.10 and 0.50 mL, respectively. The resulting attributed totals are 0.34 and 0.74 mL (Figure~\ref{fig:analytical}c). Site water usage effectiveness varies widely in practice with climate, season, and cooling technology \cite{mytton2021,siddik2021}, which makes this distinction instructionally relevant.

The values are authored assumptions, but the distinction follows directly from the separate water terms. Equal energy does not imply equal attributed water when compatible water intensities differ. Nor does the example imply that later execution generally has higher or lower water demand. It identifies information that a timing comparison would require.

\subsection{Incomplete Histories and Underdetermined Comparisons}

\begin{figure}[!t]\centering
\includegraphics[width=\textwidth]{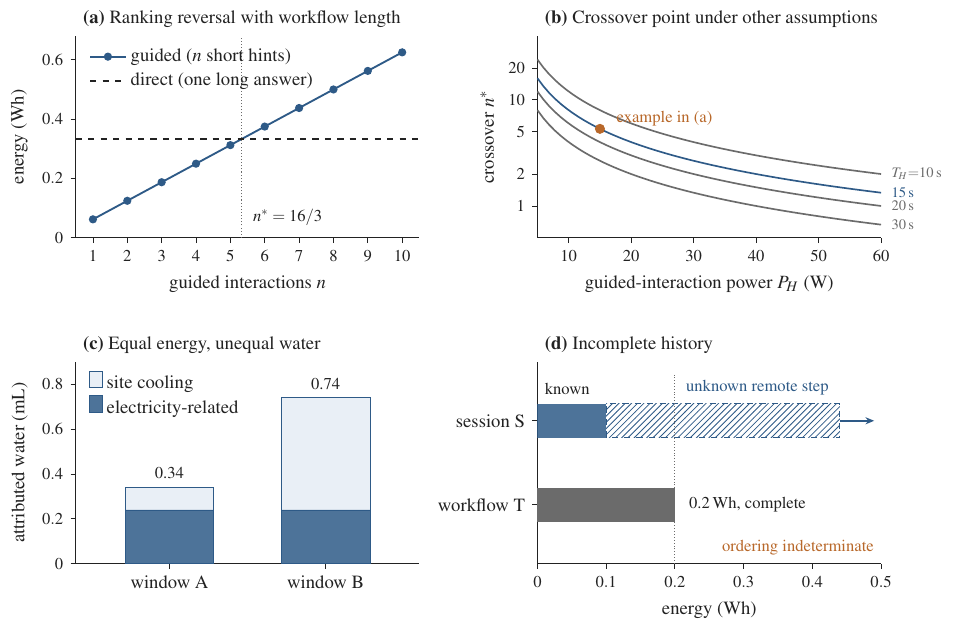}
\caption{Analytical cases computed from authored teaching assumptions, not measured inference data. (a) Ranking reversal between a direct workflow ($P_D=60$~W, $T_D=20$~s) and $n$ guided interactions ($P_H=15$~W, $T_H=15$~s); the ordering reverses at $n^*=16/3$. (b) Crossover $n^*$ (contours) for the same direct profile across guided-interaction power and duration; the star marks the example in (a). (c) Two serving windows with identical IT energy and grid water intensity but different site WUE yield different attributed water. (d) A session with an unknown component has only a lower bound and cannot be ranked against a complete 0.2~Wh workflow.\label{fig:analytical}}\end{figure}

The third case contains a known local-run energy of 0.1 Wh and an unknown remote interaction. The complete-session energy is at least 0.1 Wh under the nonnegativity assumption, but cannot be compared conclusively with a complete workflow of, for example, 0.2 Wh without further information. The unknown component could be smaller or larger than 0.1 Wh (Figure~\ref{fig:analytical}d).

These cases establish three analytical properties of the instructional model: resource ordering can reverse with workflow length; energy and water can imply different comparisons; and incomplete information can prevent an ordering. They validate the internal logic of the representations rather than the realism of selected profiles. The same parameter values are used in the prototype's comparison challenges C1--C3, and automated tests confirm that the scenario engine reproduces each value and classifies each case into the intended state. Table~\ref{tab:scenarios} summarizes the interpretations relevant to instruction.

\begin{table}[!t]\caption{Analytical cases and their instructional interpretations. Numeric profiles are authored assumptions.\label{tab:scenarios}}\centering\footnotesize\begin{tabularx}{\textwidth}{>{\raggedright\arraybackslash}p{.22\textwidth}XX}\toprule
\textbf{Case} & \textbf{Analytical result} & \textbf{Instructional interpretation}\\
\midrule
Workflow length & Direct: 0.333 Wh; four guided interactions: 0.250 Wh; six: 0.375 Wh & A resource ordering can reverse as a workflow expands\\
Water boundary & Equal IT energy; attributed water: 0.34 versus 0.74 mL & Equal energy does not establish equal water\\
Missing component & Known subtotal: 0.1 Wh; total remains unknown & Incomplete histories may not support a ranking\\
\bottomrule\end{tabularx}\end{table}

\section{Empirical Validation Framework}\label{sec:validation}

\subsection{Constructs and Formative Evaluation}

The framework's educational mechanisms require validation beyond analytical consistency. Resource literacy, help-seeking, usability, and programming learning are therefore defined as separate constructs. Resource literacy concerns interpretation and justification; help-seeking concerns actions and reasons; usability concerns interaction success; programming learning concerns independent application. Concern and confidence are supplementary self-reports rather than substitutes for these constructs.

A first formative episode should combine technical verification with structured expert appraisal. Experts in introductory programming, learning/HCI design, and energy-and-water accounting independently inspect selected missions and scenarios. They assess instructional appropriateness, technical accuracy, possible misconceptions, accessibility, and the adequacy of the evidence plan. Written explanations and disagreements are retained alongside any ratings. A revision ledger links each issue to a design decision and artifact version.

Student walkthroughs then examine whether the intended processes occur. Novice learners complete selected missions and unfamiliar interpretation tasks under a fixed script. Code histories, decisions, explanations, observed breakdowns, and retrospective interviews provide complementary evidence. Researcher assistance is recorded because it can affect both task success and interpretation. Think-aloud data \cite{ericsson1993} are treated as diagnostic observations, rather than naturalistic estimates of prompt frequency or completion time.

Analysis can begin with categories from the conjecture map and extend them to unanticipated responses. Avoidance of necessary help, guilt, distrust, and ignoring the resource information are relevant negative cases. The report should specify the coding unit, independent coding procedure, reconciliation, and version changes. Recruitment and consent require an institution-specific ethics determination before data collection.

\subsection{Instrument Development and Comparative Study}

The candidate resource-literacy tasks in Appendix~\ref{app:assessment} cover units, complete histories, energy-water separation, missingness, and execution timing. Three of them (A2--A4) are already embodied in the prototype's comparison challenges, which add two further distinctions: boundary compatibility (C4) and a trade-off in an unfamiliar context (C5). Each challenge requires a written justification, which the analysis tables export with an empty rubric column for independent coding. Content review and cognitive interviewing \cite{willis2005} are needed before the tasks can be treated as an instrument. A correct choice without a relevant explanation may reflect guessing; the proposed 0-3 explanation rubric therefore evaluates reasoning as well as selection. Item-specific anchors and rater disagreement should be reported.

A later controlled study can hold programming tasks, assistance options, pacing, and completion criteria constant while varying the resource explanations and reasoning activities. The prototype implements this contrast directly: in the control condition the resource ledger and estimates are hidden, the resource briefings are replaced by rover-engineering briefings of the same length and format, and each comparison challenge is replaced by a time-matched reflection prompt about the programming task, while missions, checks, the coach, and the model are unchanged. This comparison estimates the contribution of the resource-learning component. Changing the model or hint availability at the same time would yield a different, broader contrast.

The primary confirmatory analysis can compare post-intervention resource-literacy performance between conditions while adjusting for baseline literacy and prespecified prior experience. Randomization and analysis must preserve the participant or classroom as the experimental unit, with repeated missions treated as nested observations \cite{raudenbush2002}. The score, model, exclusion rules, missing-data treatment, meaningful effect, and multiplicity strategy should be preregistered \cite{nosek2018}. A sample-size justification follows from those decisions rather than an arbitrary target.

Independent programming performance should be assessed using a new task, with a delayed assessment if retention is a claim. A noninferiority conclusion requires a substantively justified margin and adequate power \cite{walker2011}; a nonsignificant difference alone does not show that learning was preserved. The logs support prespecified behavioural outcomes on the same model and hardware in both conditions: the distribution of hint levels, error explanations, solutions requested and declined at the confirmation step, follow-ups, regenerations, prompt and output tokens, and measured GPU energy per mission. Because fewer requests can reflect either deliberate help-seeking or disengagement, these outcomes must be interpreted together with mission success and the independent task. Other help-seeking patterns and subgroup differences remain exploratory unless separately specified and supported. Measured GPU energy per request can show whether behaviour in the study changes the energy of local assistance; it cannot establish a change in data-center load, which would require a separate deployment measurement study.

\section{Discussion}

\subsection{Educational Contribution and Generalizability}

The framework reframes resource awareness as instruction in evidence-based decision-making during programming. Its distinctive feature is the connection among an assistance choice, a complete workflow, a technically interpretable resource representation, and an independently assessable outcome. This connection can be realized in games, notebooks, or conventional programming environments; Scout provides a concrete setting in which to specify it. It complements curricular efforts that address sustainability through software-engineering content \cite{peters2024} by locating resource reasoning in a routine learning decision.

The framework also makes disagreement with the intended design empirically useful. A learner who requests more assistance but later performs better independently is not automatically an adverse case. A learner who reduces prompting while failing to understand the program is not automatically a success. Keeping the constructs separate permits a more informative account of educational value than a single resource score.

\subsection{Token Awareness as a Transferable Literacy}

Scout records prompt and output tokens for every coach request and displays them beside the energy and water records. This element may transfer beyond the game because tokens are the shared unit of commercial pricing, rate limits, context windows, and local serving logs. A student who has learned that a request includes all the code and conversation history it carries, that a regenerated answer repeats the full computation, and that a complete solution produces far more output than a targeted hint has acquired reasoning that applies to cloud assistants, institutional services, and agentic tools alike. Corresponding practices include sending the relevant function rather than an entire file, asking a narrower question before requesting a solution, and starting a new thread when earlier context is no longer needed. Whether students adopt these practices, and whether they do so without avoiding necessary help, is an empirical question that the logged behavioural outcomes in Section~\ref{sec:validation} are designed to answer.

Token awareness also has limits that the framework makes explicit. Token counts depend on the tokenizer, so counts from different models are not directly comparable; the energy of a token depends on model size, hardware, batching, and caching \cite{wilkins2024,samsi2023}; and hidden reasoning tokens can make the visible exchange an underestimate of the computation \cite{openai2025reasoning}. Scout therefore presents a token count as an evidence-labelled proxy that must be converted through a declared boundary, consistent with principle P4. For institutions, token logs offer a practical basis for course-level budgeting and for comparing assistance policies, provided that they are interpreted alongside learning outcomes rather than treated as a target to minimize.

\subsection{Scenarios, Measurement, and Institutional Responsibility}

Authored scenarios support controlled exposure, reproducible comparisons, and deliberate counterexamples. They are particularly useful when live telemetry is unavailable or when an instructional contrast requires conditions that would otherwise be difficult to reproduce. Their ecological validity depends on subsequent work with relevant learners and deployments.

Live measurement can be introduced as a separate mode with appropriate source and boundary labels. It should not erase uncertainty or be interpreted as automatically measuring a student's marginal effect. Institutions and providers retain responsibility for model serving, infrastructure efficiency, access, and reporting \cite{wu2022,shehabi2024}. Student resource literacy complements those responsibilities rather than transferring them to individual learners.

\subsection{Limitations and Scope of Inference}

The present work is a theory-informed design and analytical contribution. The literature synthesis is selective, the prototype has been verified with automated tests but not evaluated with learners, and no participant outcomes are reported. Its execution limits protect against accidental errors but do not isolate untrusted code, so classroom deployment on shared machines requires the specified isolated-execution service; the answer guard is a heuristic whose effectiveness with real model output remains to be measured. The GPU meter covers the graphics boards only, and the power reported by \texttt{nvidia-smi} can be averaged or sampled over windows comparable to a short request \cite{yang2023nvsmi}; its accuracy for individual requests should therefore be checked against an external meter and repeated batches before per-request values are interpreted. The optional large-cloud assistant offered in non-study mode is a labelled simulation rather than a measured service. The analytical examples establish mathematical relationships within declared assumptions. They do not validate hardware profiles, model quality, water coefficients, or learning effects.

Future studies must address construct validity, novelty effects, uneven prior knowledge, outside-tool use, and the possibility that resource displays discourage necessary help. Scenario sets should include reversals, unknowns, and cases where additional assistance is educationally justified. Generalizability will require different tasks and institutions, while claims about shared-server performance require measured concurrent workloads. These limitations delimit the conclusions without reducing the framework to a claim of effectiveness made in advance of evidence.

\section{Conclusions}

Resource-aware LLM-assisted programming education requires students to understand what a resource comparison represents and how it relates to a learning decision. Scout connects that objective with staged Python activity, actionable assistance, workflow visibility, epistemic transparency, learner agency, and transfer-oriented assessment. The formal scenario model supplies compatible energy-and-water definitions, explicit missingness, and comparison rules that accommodate uncertainty and trade-offs. Analytical cases establish why prompt counts, equal energy, and incomplete histories cannot support simplistic rankings. Because programming is both the largest and one of the most token-intensive uses of coding assistants, and because introductory courses are where help-seeking habits form, token awareness, measured locally and interpreted through declared boundaries, belongs in programming education alongside the complexity of the programs that students write. The framework's next validation step is a documented cycle of expert appraisal, student observation, and revision, followed by a controlled assessment of resource literacy and independent programming. Its educational value will depend on those outcomes, rather than on reduced prompting or visual sophistication alone.

%% file: appendix.tex
\begin{appendices}
\section{Candidate Resource-Literacy Assessment}\label{app:assessment}

The following tasks operationalize the working construct and require content review and cognitive interviewing before use as a validated instrument. They are intended for unfamiliar contexts rather than repetition of the teaching examples. Tasks A2, A3, and A4 correspond to prototype challenges C1, C2, and C3; A1, A5, and A6 are not yet implemented in the game.

A1. A request has a recorded duration but no power reading or calibrated power profile. What additional information is required to estimate its energy? The target distinction is the difference between elapsed time and energy.

A2. Two workflows complete the same task, one with two long responses and another with five short hints. What evidence is needed to compare them? The target distinction is the difference between visible interaction counts and complete workload.

A3. Two serving windows have equal IT energy but different site-water intensities. Can their attributed cooling water differ? The target distinction is the independence of energy equality and water equality under different intensities.

A4. A session includes one known-energy run and one unknown-energy remote interaction. What does the known-component sum establish? The target distinction is the difference between a subtotal and a complete total.

A5. A student requests an additional conceptual explanation and subsequently solves a new problem independently. How should the request be interpreted? The target distinction is the separation of resource use from educational adequacy.

A6. An interface waits before displaying an answer that has already been generated. What evidence would show whether computation was rescheduled? The target distinction is execution timing versus display timing.

Explanations can be scored from 0 to 3 with item-specific anchors: 0 identifies the target misconception; 1 indicates a partly correct choice without a relevant reason; 2 identifies the distinction but omits an important boundary or condition; and 3 identifies the distinction and the relevant condition or missing evidence. Raters should remain blind to condition where feasible and report agreement before adjudication. Aggregation into a total score requires additional evidence about the interpretation and reliability of that score.

\end{appendices}